\documentclass[10pt]{article}
\usepackage[margin=0.82in]{geometry}
\usepackage{graphicx}
\usepackage{booktabs}
\usepackage{amsmath,amssymb}
\usepackage{microtype}
\usepackage[numbers,sort&compress]{natbib}
\usepackage[hidelinks]{hyperref}
\usepackage{enumitem}
\usepackage{caption}
\usepackage{xcolor}
\setlist{nosep,leftmargin=*}
\title{Path-Conditioned Attention Mobility: From Depth to Execution Hubs in SegFormer}
\author{Aleksandr Druzhinin\\
Independent Researcher\\
\texttt{sdruzhinin1992@gmail.com}}
\date{}
\begin{document}
\maketitle
\begin{abstract}
Transformer compression commonly reduces attention cost by pruning heads, dropping layers, or sharing computations across depth. These approaches often couple two distinct quantities: the amount of attention capacity that must be retained and the number of sequential depths at which that capacity must be executed. We study whether useful attention functions in a hierarchical vision transformer are instead functionally mobile across depth.

Using SegFormer-B5 for human segmentation, we find that some task-relevant attention heads can be relocated to nearby depths, but this mobility is strongly conditioned on their computational context. Head-only relocation is insufficient, while adaptation of the surrounding representation boundary substantially improves functional preservation. More importantly, replacing donor-specific normalization and spatial-reduction preprocessing with shared or recipient-side paths causes large degradation.

We exploit this observation to reorganize retained attention from depth into width. Starting from a fixed parent with 120 retained Stage-3 heads distributed across 40 attention execution depths, we consolidate the same 120-head budget into 20 execution hubs. Under a layout-matched three-seed recovery control, K20 reaches $0.960686\pm0.000059$ mIoU compared with $0.960400\pm0.000074$ for the corresponding State120 control. The paired point difference is $+0.000285$, while a paired image-bootstrap confidence interval includes zero. We therefore interpret the result as no measurable quality penalty rather than statistical superiority.

To realize the structural transformation efficiently, we introduce FOLDED\_VEC, a deployment-only reparameterization that preserves donor-specific paths while vectorizing their execution. Relative to the same 120-head State120 parent, the final K20 deployment reduces end-to-end NVIDIA L4 latency by approximately 5.28\%, 3.15\%, and 3.58\% at batch sizes 1, 4, and 8, respectively. Across the full 572-image external development observer and three fixed seeds, folding changes mean mIoU by only $-5.81\times10^{-7}$.

Our results suggest that attention redundancy is not purely a question of removable capacity. Useful attention functions can be locally mobile across depth, provided that the computational path defining their function is preserved.
\end{abstract}
\section{Introduction}
Multi-head self-attention is a major source of both representational capacity and computational cost in modern Transformers. At the same time, a substantial body of work has shown that not all attention capacity learned during training is required at inference. Attention heads can often be removed with limited degradation, motivating structured head pruning and related compression techniques. In vision transformers, this observation has led to methods that prune attention, reduce network depth, reuse attention computations, merge tokens, or dynamically allocate compute.

Most such approaches, however, couple two distinct quantities: attention capacity and attention execution depth. Head pruning asks how many attention functions can be removed. Layer pruning asks which sequential transformations can be skipped. Cross-layer sharing asks whether related layers can reuse attention computations. These formulations leave a different question comparatively underexplored:

\begin{quote}
\emph{If an attention function is useful enough to retain, must it still be executed at the depth at which it was originally learned?}
\end{quote}

The distinction matters because redundancy across depth does not necessarily imply interchangeability across depth. Cross-layer sharing methods already suggest that attention computations in neighboring Transformer layers can be correlated. More recent methods transfer, align, or approximate attention information between neighboring layers. Their need for alignment, correction, or re-initialization, however, also suggests that treating attention functions from different depths as freely interchangeable is insufficient.

We investigate this problem in a hierarchical vision transformer, where cross-depth relocation is further complicated by depth-dependent feature representations and spatial-reduction preprocessing. Rather than beginning with a new compression architecture, we first ask a functional question: does a pretrained attention head preserve task-relevant behavior when executed at a nearby depth?

In SegFormer-B5, a systematic Stage-3 mobility study reveals that some task-relevant heads are indeed locally mobile. However, this mobility is conditional rather than universal. Relocating attention parameters alone is insufficient: in our controlled adaptation study, jointly adapting the relocated attention and its surrounding representation boundary is consistently more effective than head-only adaptation. More importantly, replacing donor-specific normalization and spatial-reduction preprocessing with shared or recipient-side paths causes substantial degradation.

These observations motivate what we call \emph{path-conditioned functional attention mobility}: useful attention can move across nearby depths, but its function depends on the computational path through which its input representation is transformed.

This perspective suggests a different form of structural compression. Instead of asking only which attention functions can be deleted, we ask whether retained attention functions can be reorganized from depth into width. We first reduce redundant attention capacity and freeze a parent model with 120 retained Stage-3 heads distributed across 40 attention execution depths. We then construct attention execution hubs that gather retained heads originating from multiple nearby depths while preserving their donor-specific preprocessing paths.

The resulting K20 configuration retains exactly the same 120-head budget but requires only 20 attention execution depths. Thus, unlike conventional pruning comparisons, the reduction in attention depth is not confounded by further removal of retained attention capacity.

This fixed-capacity comparison provides the central empirical test of the paper. We construct layout-matched State120 controls that preserve the original 40-depth organization while receiving the same recovery opportunity and closely matched trainable components. Across three fixed seeds, K20 reaches external-development mIoU $0.960686\pm0.000059$, compared with $0.960400\pm0.000074$ for its layout-matched State120 control. K20 is higher in all three paired runs. The paired point difference is $+0.000285$, while a 20,000-sample paired image bootstrap yields a confidence interval that includes zero. We therefore interpret the result as no measurable quality penalty, not statistical superiority.

Reducing logical execution depth does not automatically produce proportional hardware acceleration. State120 and K20 perform essentially the same retained-head arithmetic, and preserving donor-specific preprocessing introduces heterogeneous paths that fragment GPU execution. We therefore separate functional consolidation from deployment realization.

Our FOLDED\_VEC transformation algebraically folds compatible affine normalization terms into subsequent donor-specific operations and vectorizes their execution without forcing donor paths to share parameters. Relative to State120, the final folded K20 implementation reduces end-to-end NVIDIA L4 latency by 5.28\%, 3.15\%, and 3.58\% at batch sizes 1, 4, and 8. Across all 572 external-development images and three fixed seeds, the mean mIoU change introduced by folding is only $-5.81\times10^{-7}$.

Our results therefore suggest that attention redundancy has at least two distinct dimensions. Capacity redundancy determines which attention functions can be removed. Execution-depth redundancy determines how many sequential attention locations are required for the functions that remain. The latter cannot be exploited by arbitrary cross-layer sharing because the computational path of a relocated function remains important.

The contributions of this work are:
\begin{enumerate}
\item \textbf{Path-conditioned functional attention mobility.} We systematically show that some task-relevant attention functions in SegFormer-B5 are locally mobile across depth, without claiming universal portability.
\item \textbf{Mechanistic evidence for path dependence.} Joint relocation adaptation substantially outperforms head-only adaptation, and donor-specific normalization and spatial-reduction preprocessing are critical to preserving relocated attention.
\item \textbf{Fixed-capacity depth-to-width consolidation.} We introduce attention execution hubs and show that the same 120 retained heads can be reorganized from 40 to 20 attention execution depths without a measurable quality penalty under a matched control.
\item \textbf{A hardware-realizable deployment transformation.} FOLDED\_VEC preserves heterogeneous donor paths while reducing execution fragmentation, converting the structural reorganization into a measurable latency reduction relative to the same 120-head parent.
\end{enumerate}

\begin{figure*}[t]
\centering
\includegraphics[width=\textwidth]{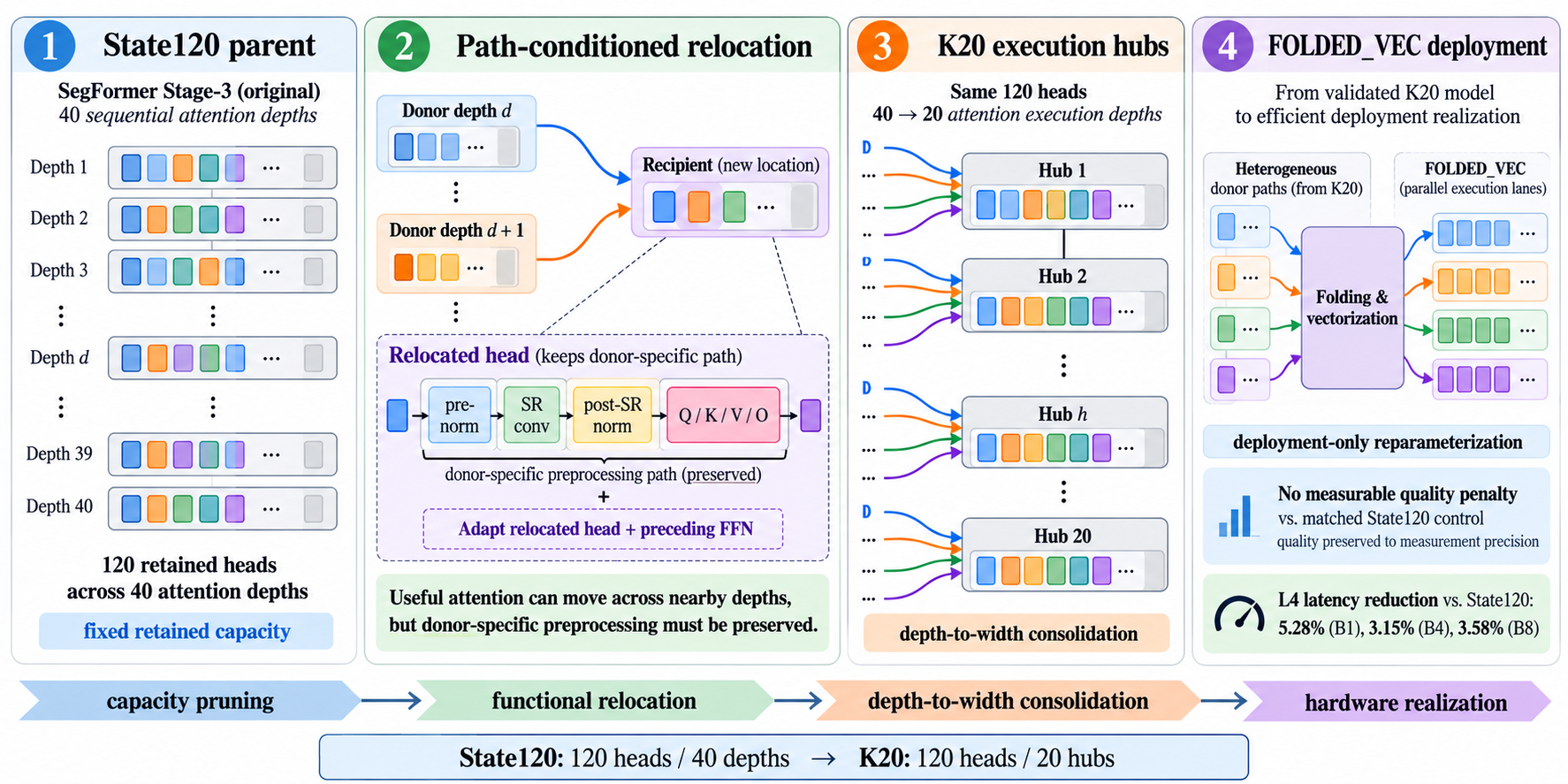}
\caption{Overview of path-conditioned attention mobility and execution-hub consolidation. Starting from the fixed 120-head State120 parent, selected attention functions are relocated to nearby execution depths while retaining donor-specific preprocessing. Multiple heterogeneous donor functions are then executed at common attention hubs, reducing Stage-3 attention execution depths from 40 to 20 without changing the retained 120-head budget. FOLDED\_VEC is an inference-only deployment reparameterization that folds and vectorizes compatible donor-specific operations without parameter sharing.}
\label{fig:overview}
\end{figure*}

\section{Related Work}

\subsection{Efficient SegFormer and Token Reduction}

SegFormer combines a hierarchical Transformer encoder with a lightweight MLP decoder and has become a widely used architecture for semantic segmentation \citep{xie2021segformer}. Its larger variants achieve strong segmentation accuracy but incur substantial computational cost, motivating efficiency improvements at several structural levels.

One line of work reduces the amount of spatial information processed by the Transformer. Token pruning and token merging decrease the number of tokens propagated through subsequent layers rather than modifying the attention functions themselves. SegFormer++ adapts token-merging strategies specifically to SegFormer for high-resolution semantic segmentation and evaluates several merging schemes across segmentation and human-pose tasks \citep{kienzle2024segformerpp}. Without model retraining, the authors report a 61\% inference acceleration on Cityscapes while maintaining mIoU.

SegFormer++ is complementary to our approach. Token merging compresses the spatial/token dimension by reducing the amount of data processed by subsequent Transformer blocks. Our

method instead compresses the attention execution-depth dimension: the token sequence remains unchanged, while retained attention functions are reorganized across depth. These two axes are not mutually exclusive and could in principle be combined.

This distinction is useful because efficiency in hierarchical vision transformers can arise from several independent sources: fewer tokens, fewer attention heads, fewer channels, or fewer sequential execution points. Our work focuses specifically on the last of these while controlling for retained attention capacity.

\subsection{Structured Attention and Depth Pruning}

Redundancy in multi-head attention has been studied extensively. Michel et al. showed that many attention heads in trained Transformer models can be removed with limited performance degradation, establishing head-level redundancy as an important target for structured compression \citep{michel2019sixteen}. Subsequent vision-transformer pruning methods have extended structured compression across attention heads, channels, MLP dimensions, tokens, and network depth.

Recent work has also studied structured pruning directly in SegFormer. DynaSegFormer uses input-dependent dynamic gating to prune SegFormer computation and combines it with knowledge distillation \citep{bai2022dynasegformer}. StaticSegFormer \citep{bartels2026staticsegformer} explicitly contrasts a static structured-pruning design with dynamic pruning and emphasizes that theoretical FLOP reduction is not sufficient: static architectures may map substantially better to GPU execution than dynamic schemes. StaticSegFormer reports frame-rate improvements of up to 34\% on Cityscapes without mIoU degradation in its tested setting. This result makes an important distinction for our work: we do not claim that pruning attention heads in SegFormer is itself novel, nor do we position our method primarily as a stronger head-pruning algorithm.

Instead, we separate attention capacity from attention execution depth. Conventional head pruning asks which attention functions can be removed. After this decision has been made, we ask a different question: must the attention functions that remain still be executed at all of their original sequential depths?

Depth-oriented ViT compression provides a related perspective. DC-ViT \citep{zhang2024dcvit} is a few-shot ViT compression framework that avoids dropping complete Transformer blocks and instead selectively removes attention modules while retaining and reusing parts of the corresponding MLP computation, thereby producing a dense set of compression operating points. HetDPT \citep{su2026hetdpt} more recently analyzes heterogeneity across different Transformer component types, including attention, FFN, and activation layers, through gradient disparity and recovery asymmetry, motivating heterogeneity-aware depth pruning rather than treating all components as equivalent.

Our method differs from these approaches in that useful attention from eliminated execution depths is not discarded. The retained attention functions are relocated and consolidated into other execution points, allowing depth to be reduced without necessarily reducing the retained attention budget.

\subsection{Cross-Layer Attention Sharing and Reuse}

A second line of work exploits redundancy between attention computations at different depths. LazyFormer \citep{ying2021lazyformer} reduces repeated self-attention computation by evaluating an attention distribution once and reusing it over multiple subsequent Transformer layers. In vision transformers, PSViT \citep{chen2021psvit} similarly observes strong correlation between attention maps of neighboring layers and introduces attention sharing together with token pooling.

These works demonstrate that attention calculations across adjacent depths can contain substantial redundancy. However, they primarily exploit this redundancy by sharing the result of an attention computation. Our setting is different: retained donor heads preserve their own learned Q/K/V/O parameters and may coexist at a common execution hub without being assumed to implement the same attention function.

LiSA \citep{mu2026lisa} provides a particularly relevant modern comparison. It studies cross-layer attention redundancy in pretrained large language models and finds that directly sharing attention weights between adjacent layers is ineffective unless attention heads are first aligned. LiSA therefore learns lightweight feed-forward alignment functions and low-rank corrections to approximate differences between layer-wise attention weights.

This finding is consistent with an important premise of our work: redundancy across depth does not imply that layers are freely interchangeable. We reach this conclusion from a different direction. Rather than approximating several layers with a shared attention computation, we relocate individual pretrained attention functions and experimentally determine which parts of their original computational context must remain intact.

In SegFormer, this issue is especially important because attention is preceded by hierarchical, depth-dependent spatial-reduction preprocessing. Our ablations show that replacing donor-specific preprocessing with a shared or recipient-side path can cause large performance losses. Thus, the effective function of a head cannot be identified solely with its Q/K/V/O matrices; it is partly defined by the path through which the representation reaches those matrices.

\subsection{Cross-Layer Transfer and Depth-to-Width Reorganization}

Recent compression methods have begun to preserve useful information from layers whose sequential depth is removed, rather than treating depth pruning as pure deletion.

Pangu Light \citep{chen2025pangulight} is one of the closest precedents. Its Cross-Layer Attention Pruning (CLAP) transfers selected attention parameters from pruned LLM layers into neighboring retained layers as part of a weight re-initialization strategy for aggressive structured pruning. The framework jointly targets model width, depth, attention heads, and normalization components and also employs inference-oriented parameter absorption. Consequently, neither cross-layer transfer of pretrained attention parameters nor normalization folding should be considered novel in isolation.

Our objective differs from CLAP in an important way. Cross-layer transfer in Pangu Light is introduced primarily as a recovery mechanism after structured pruning. We instead treat functional relocation itself as the phenomenon to be characterized. We measure head mobility across the complete Stage-3 attention population, compare relocation against appropriate controls, study the adaptation required after relocation, and identify the donor-specific preprocessing path as a critical condition for preserving function. We then use these observations to design a new execution topology rather than only to initialize a pruned network.

FlattenGPT \citep{xu2026flattengpt} is an especially close depth-to-width precedent. It flattens adjacent Transformer blocks into a shallower, wider structure rather than discarding an entire removed block, and then prunes channel-level redundancy to recover a compact model. Thus, depth-to-width reorganization itself is not unique to our work. Our execution hubs differ in granularity and architectural constraints: we relocate individual retained attention heads rather than flatten entire blocks; the FFNs remain at their original depths; donor-specific pre-normalization, spatial-reduction, and post-reduction normalization paths are preserved in the hierarchical SegFormer encoder; the frozen K20 topology uses local relocation with a maximum displacement of four Stage-3 depths; and the primary State120--K20 comparison keeps the 120-head budget fixed under a layout-matched control.

This fixed-capacity formulation is central to our analysis. In our primary comparison, State120 and K20 contain the same 120 retained Stage-3 heads. State120 executes them across 40 original attention depths, whereas K20 consolidates them into 20 execution hubs. The comparison therefore isolates attention execution depth from additional attention-capacity removal.

\subsection{Functional Compression and Hardware Realization}

Structural compression and hardware acceleration are related but distinct objectives. A reduction in theoretical depth, parameter count, or FLOPs does not guarantee a proportional decrease in wall-clock latency. StaticSegFormer explicitly demonstrates this issue by showing that models with comparable mIoU and FLOPs can exhibit substantially different GPU throughput depending on whether their pruning structure maps efficiently to hardware.

We observe the same distinction in our setting. Preserving donor-specific preprocessing is important for functional correctness, but naively executing these heterogeneous donor paths introduces many small operations and limits the latency benefit of attention consolidation. We therefore separate functional compression from deployment realization.

After the hub architecture has been validated, FOLDED\_VEC algebraically folds compatible affine normalization terms into subsequent donor-specific projections and spatial-reduction operations and vectorizes their execution. Importantly, this does not replace the donor paths by a shared approximation; it changes how the same validated heterogeneous computations are physically executed.

The resulting hardware optimization should therefore be interpreted as a systems contribution built on top of the functional result, rather than as additional evidence of model redundancy.

\subsection{Positioning of This Work}

Prior work establishes that attention heads can be pruned, tokens can be merged, attention maps can be reused between layers, depth can be reduced without pure block deletion, and selected attention information can be transferred across neighboring layers. The unresolved question we study is more specific:

When a useful pretrained attention function is retained, how strongly is that function tied to the depth at which it was learned, and what computational context must be preserved if it is executed elsewhere?

Our experiments indicate an intermediate regime between strict depth specificity and unrestricted cross-layer interchangeability. Some attention functions are locally mobile, but their mobility is path-conditioned. In particular, donor-specific preprocessing and adaptation of the surrounding representation boundary play an important role in preserving the relocated function.

We exploit this property to reorganize attention from depth into width. Rather than decreasing the head budget further, we retain the same 120 Stage-3 heads and consolidate their execution from 40 depth locations into 20 attention hubs. This distinguishes our approach from pure head pruning, layer dropping, token merging, and attention sharing: the retained functions remain heterogeneous, but fewer sequential execution points are required to realize them.

Accordingly, the main contribution of this work is not another demonstration of attention redundancy. It is the characterization and exploitation of path-conditioned functional attention mobility in a hierarchical vision transformer.

\section{Method}
\subsection{Problem Formulation}
Consider a task-adapted hierarchical vision transformer containing attention-bearing blocks at depths $i=1,\ldots,L$. Let
\[
\mathcal{H}_i=\{h_{i,1},\ldots,h_{i,n_i}\}
\]
denote the attention heads originally executed at depth $i$. Standard head pruning reduces the total retained attention capacity
\[
N_{\mathrm{heads}}=\sum_i |\mathcal{H}_i|
\]
by deleting elements of $\mathcal{H}_i$. Depth pruning instead reduces the number of sequential Transformer blocks, usually removing all computation associated with selected depths.

We distinguish two quantities: the number of retained attention heads $N_{\mathrm{heads}}$ and the number of sequential locations at which attention is executed, $N_{\mathrm{attn\mbox{-}depths}}$. Our central question is whether reducing the second quantity necessarily requires further reducing the first.

For SegFormer-B5 Stage 3, the source model contains 200 attention heads distributed across 40 attention depths. After head thinning, our frozen parent, denoted State120, retains 120 heads while preserving the original 40-depth execution structure. We seek a transformation
\[
\{\mathcal{H}_1,\ldots,\mathcal{H}_{40}\}\rightarrow\{\mathcal{G}_1,\ldots,\mathcal{G}_K\},
\]
where each $\mathcal{G}_k$ is an attention execution hub containing heads that originally belonged to one or more donor depths, subject to
\[
\sum_{k=1}^{K}|\mathcal{G}_k|=120, \qquad K<40.
\]
Thus, the retained attention capacity is held fixed while execution depth changes.

\subsection{Head Thinning and the Frozen State120 Parent}
Before studying execution-depth redundancy, we first reduce clear head-level redundancy. Head removal is performed sequentially rather than from a single static ranking. At each step, candidate heads are screened using a low-cost proxy and then evaluated on a task-derived calibration subset. One head is removed at a time, with at least one head retained in every Stage-3 block. No recovery training is performed during this trajectory.

The trajectory is deliberately evaluated beyond mild compression. Early pruning is largely redundancy-dominated, whereas deeper pruning becomes non-monotonic and increasingly dependent on interactions with previously removed heads. We therefore do not interpret any single head count as a universal intrinsic capacity threshold.

For all subsequent relocation and hub experiments, we freeze the 120-head state obtained at the predeclared step of this trajectory. State120 is not selected because it maximizes the external observer; the highest observed external point occurs earlier, near 150 heads. State120 instead provides a fixed-capacity parent from which we can study where retained attention is executed.

\subsection{Functional Attention Mobility}
Let $h_{i,j}$ denote head $j$ originally associated with donor depth $i$. We define a local relocation
\[
M(h_{i,j}, i\rightarrow r),
\]
where the pretrained head parameters are executed at a nearby recipient depth $r$, while the remainder of the network is kept fixed.

The mobility study compares three conceptually distinct interventions: KEEP, DROP, and MOVE. DROP removes the head entirely. MOVE removes it from its original execution location but preserves its pretrained attention function at a neighboring depth. Mobility is supported when relocation preserves task function that would otherwise be lost by dropping the same head.

We apply this analysis across all 200 original Stage-3 heads and evaluate 390 directed adjacent moves, together with controls and a radius study. The resulting evidence supports local functional mobility for some task-relevant heads, but not arbitrary-distance portability or universal superiority of pretrained relocation over reinitialization. Functional mobility is therefore treated as a property of a head under a specific relocation context rather than as a permanent binary label.

\subsection{Relocation Requires Representation-Boundary Adaptation}
A head trained at donor depth $i$ does not receive the same representation when executed at recipient depth $r$. Consequently, transferring Q/K/V/O parameters alone need not preserve the original function.

We distinguish three relocation variants: RAW MOVE, HEAD\_ONLY\_FT, and HEAD+PRE\_FFN\_FT. RAW MOVE relocates the pretrained head without adaptation. HEAD\_ONLY\_FT adapts the relocated attention parameters while the surrounding network remains frozen. HEAD+PRE\_FFN\_FT additionally adapts the immediately preceding representation boundary.

The third condition is not intended to establish that a specific FFN is by itself the unique causal mechanism. Instead, it tests whether successful relocation requires adaptation beyond the head parameters alone. Across the controlled 24-head study, joint head and boundary adaptation is consistently more effective than head-only adaptation. This motivates the first part of our path-conditioned mobility hypothesis:
\[
\text{relocated function}\neq\text{attention parameters alone}.
\]

\subsection{Donor-Specific Preprocessing}
SegFormer attention introduces an additional complication because the key/value path includes spatial-reduction preprocessing. For donor depth $i$, let $P_i(x)$ denote the donor-specific preprocessing applied before the corresponding attention computation. In our implementation this path includes the relevant pre-normalization, spatial-reduction operation, and post-reduction normalization.

A donor head is therefore represented as
\[
h_{i,j}(x)=A_{i,j}\!\left(P_i(x)\right),
\]
where $A_{i,j}$ contains the retained attention projections. A naive hub could instead evaluate the donor head through the recipient preprocessing path $A_{i,j}(P_r(x))$ or force several donors to share a common path $P_{\mathrm{shared}}$. We test these alternatives explicitly.

The mechanism ablation separately shares the pre-normalization, spatial-reduction convolution, post-reduction normalization, the complete spatial-reduction path, or the recipient path. All cause degradation relative to exact donor-path execution, with especially large losses when the spatial-reduction convolution or complete recipient preprocessing is shared. We therefore impose the architectural constraint that relocated attention retains donor-specific preprocessing while preserving its retained pretrained Q/K/V/O parameters.

This yields the stronger form of the path-conditioned mobility hypothesis: an attention function may be mobile across execution depth even when the preprocessing path supporting that function is not freely interchangeable across depths.

\subsection{Attention Execution Hubs}
Given a fixed collection of retained heads, we consolidate their execution locations. An attention hub $\mathcal{G}_k$ is associated with a recipient execution depth $r_k$ and a set of donor heads
\[
\mathcal{G}_k=\{h_{i,j}:\pi(i,j)=k\},
\]
where $\pi$ maps every retained head to exactly one hub. Heads originating from different donor depths retain their donor-specific preprocessing paths; for a donor head executed at hub $k$,
\[
z_{i,j}=A_{i,j}\!\left(P_i(x_{r_k})\right).
\]
The resulting head outputs are combined at the common hub execution point rather than requiring independent sequential attention modules at every original donor depth.

The transformation therefore keeps $N_{\mathrm{heads}}=120$ fixed while changing
\[
N_{\mathrm{attn\mbox{-}depths}}: 40\rightarrow K.
\]
We construct a nested consolidation frontier $K\in\{20,16,14,12,10\}$ with bounded relocation distance and parent-to-child nested layouts. The final replicated evaluation focuses on K20, K14, and K10 as conservative, balanced, and aggressive operating points.

\subsection{Recovery After Consolidation}
Relocation changes the representation boundary at which donor attention operates. We therefore allow a short, fixed recovery stage after constructing each hub layout. Only parameters directly involved in relocation and local representation adaptation are updated: relocated retained-head Q/K/V/O parameters and the corresponding pre-hub FFN components. The rest of the compressed network remains fixed.

All final operating points use the same predetermined recovery duration and fixed seeds; no best epoch or best seed is selected from the external observer. Conceptually, recovery solves
\[
\min_{\theta_{\mathrm{move}},\theta_{\mathrm{boundary}}}\mathcal{L}_{\mathrm{task}}
\]
subject to the already fixed hub assignment $\pi$. Architecture selection and recovery are therefore separated: recovery adapts a frozen topology rather than searching for a new one.

\subsection{Layout-Matched State120 Control}
Additional training alone can improve segmentation quality, so comparing a recovered hub model only with the raw State120 parent would confound architectural consolidation with optimization. We therefore construct a layout-matched control for every final hub configuration.

For a given hub layout, the control retains the original State120 architecture with 120 heads at 40 attention depths. No attention head is relocated and no attention execution depth is removed. However, it trains the same classes of parameters that would be adapted in the corresponding hub model: the retained-head Q/K/V/O parameters associated with moved donors and the same immediately preceding FFN components. The control uses the same training data, optimizer policy, fixed recovery length, and seeds.

The resulting paired comparison isolates the architectural question of 40 original attention depths versus $K$ hubs while approximately matching the additional optimization opportunity. This is the primary causal control used for our depth-consolidation claim.

\subsection{Physical Hub Execution}
The functional hub definition does not by itself guarantee efficient GPU execution. A naive implementation could preserve donor-specific preprocessing by executing every donor path and attention head sequentially, fragmenting the workload into many small kernels and eliminating much of the latency benefit of reduced logical depth.

The physical implementation therefore groups heads assigned to the same hub into parallel operations. Donor-specific preprocessing is retained, while compatible Q/K/V representations are assembled into a wider attention computation and executed through parallel scaled dot-product attention. Output projections are correspondingly combined at the hub level. Thus, functional relocation and parallel physical realization are treated as separate transformations.

\subsection{FOLDED\_VEC Deployment Reparameterization}
Even after physical hub fusion, donor-specific preprocessing still contains multiple affine normalization and projection operations. We therefore introduce an inference-only transformation, FOLDED\_VEC, that preserves donor-specific parameters while reducing execution fragmentation.

Consider affine normalization
\[
\operatorname{LN}_{\gamma,\beta}(x)=\gamma\odot\operatorname{LN}_{0}(x)+\beta.
\]
For a subsequent linear projection $y=Wu+b$, substituting the affine-normalized representation gives
\[
y=W(\gamma\odot u+\beta)+b=W'u+b',
\]
with
\[
W'=W\operatorname{diag}(\gamma), \qquad b'=W\beta+b.
\]
The same principle applies to compatible donor-specific convolutional projections. FOLDED\_VEC precomputes the folded parameters and stacks compatible donor operations into wider vectorized kernels.

Crucially, folding does not impose $P_i=P_j$. Each donor retains its own learned parameters; only their physical execution is reorganized. FOLDED\_VEC therefore maps heterogeneous donor functions to vectorized heterogeneous execution rather than to a shared approximation. It is applied only after training and architecture selection and introduces no additional optimization or architecture search.

Because floating-point execution order changes, we do not assume bitwise identity between the physical K20 model and FOLDED\_VEC. Deployment equivalence is evaluated empirically.

\subsection{Complete Compression Pipeline}
The complete procedure is:
\begin{enumerate}
\item Adapt the source SegFormer-B5 to the downstream segmentation task.
\item Thin Stage-3 heads sequentially to obtain a fixed retained-head parent.
\item Freeze the predeclared State120 configuration.
\item Probe local functional mobility of pretrained heads across nearby depths.
\item Determine the adaptation required for relocation.
\item Identify donor-specific preprocessing as an architectural constraint.
\item Construct nested attention-hub layouts while keeping the 120-head budget fixed.
\item Apply a short, fixed recovery to each frozen hub topology.
\item Compare each hub configuration against its layout-matched State120 control.
\item Convert the selected hub architecture to a parallel physical implementation.
\item Apply FOLDED\_VEC as an inference-only vectorizing reparameterization.
\end{enumerate}
The conceptual separation is therefore
\[
\text{capacity pruning}\rightarrow\text{functional relocation}\rightarrow\text{depth-to-width consolidation}\rightarrow\text{hardware realization}.
\]

\section{Experimental Protocol}
\subsection{Dataset and Task}
We evaluate the proposed attention-mobility and consolidation framework on binary human semantic segmentation using the Supervisely Persons Dataset. After preprocessing, the dataset contains 5,711 image--mask pairs. We use a fixed split containing 5,139 training images and 572 images used as an external development observer. All images are processed at a resolution of $512\times512$.

Training augmentation consists of longest-side resizing and padding, random cropping, horizontal flipping, small shift/scale/rotation perturbations, and brightness/contrast augmentation. Calibration and external evaluation use deterministic resizing, padding, and center cropping.

The 572-image split requires an explicit qualification. It was repeatedly inspected during the broader research process and was historically used for model development. We therefore do not treat it as an untouched test set. Throughout this paper, we refer to it as the \emph{external development observer}.

For the head-thinning, relocation, hub-layout, and final recovery experiments reported here, architecture-selection rules, relocation constraints, operating-point definitions, recovery duration, fixed endpoints, and final seeds were determined without selecting on this external observer. The external split is used to report the behavior of already frozen trajectories and operating points rather than to choose the best seed, recovery epoch, individual relocation, or final hub configuration.

\subsection{Source Model}
Our source architecture is SegFormer-B5 adapted to binary human segmentation. The model is initialized from the ADE20K-pretrained Hugging Face checkpoint \texttt{nvidia/segformer-b5-finetuned-ade-640-640}, with the original prediction layer replaced by a two-class segmentation head.

The selected task-adapted source checkpoint is taken from epoch 108. Its Stage 3 contains 40 Transformer blocks and 200 attention heads. This checkpoint defines the initial representation and attention population from which the head-thinning trajectory and all subsequent relocation experiments are derived. The central architectural comparisons are not made against the uncompressed source model; recovered hub configurations are compared with layout-matched State120 controls that retain the same attention-head budget.

\subsection{Source-Model Training}
The source SegFormer model is optimized using AdamW with an initial learning rate of $2\times10^{-6}$, weight decay $10^{-3}$, label smoothing 0.03, effective batch size 16, cosine learning-rate decay, and gradient clipping. This checkpoint predates the final attention-hub study and provides the frozen starting point for the Stage-3 head-thinning trajectory.

\subsection{Train-Derived Head Thinning}
The head-thinning trajectory operates only on Stage 3 of SegFormer-B5, which contains 40 Transformer blocks and 200 attention heads in the source model. Candidate selection follows a sequential two-stage procedure:
\[
\text{PROXY32}\rightarrow\text{exact rotating TASK514}\rightarrow\text{remove one head}.
\]
At least one attention head is retained in every Stage-3 block. Head removals are performed one at a time, and no recovery training is applied between individual removals.

The training pool is repartitioned during structural decision intervals into a 4,625-image fitting subset and a 514-image calibration subset. Calibration subsets are generated from the training pool using deterministic but varying splits. The 572-image external observer does not participate in candidate ranking or head-removal acceptance. The thinning trajectory is continued to the predeclared 120-head parent rather than terminated based on external-development performance.

\subsection{Mobility and Relocation Experiments}
The functional mobility study covers all 200 original Stage-3 heads and evaluates 390 directed adjacent relocations. During the basic mobility probe, the network remains frozen: no gradient-based adaptation is used to decide whether MOVE preserves useful task behavior relative to DROP.

Three interventions are considered: KEEP, DROP, and MOVE. KEEP preserves the original head at its native depth. DROP removes the head. MOVE removes the head from its native execution position while executing the pretrained attention function at a nearby depth.

The mobility experiment also includes a SegFormer-native reinitialization control. This prevents a stronger interpretation that pretrained MOVE universally outperforms a newly initialized replacement. The supported claim is restricted to local functional mobility for some task-relevant heads.

To study adaptation after relocation, we evaluate 24 heads under RAW MOVE, HEAD\_ONLY\_FT, and HEAD+PRE\_FFN\_FT. These experiments use three fixed seeds and a predetermined epoch-3 endpoint. No best recovery epoch is selected using the external development observer.

\subsection{Donor-Path Mechanism Ablation}
Using a packed K10 topology, we compare exact donor-specific execution with variants that replace individual donor components by shared or recipient-side alternatives. The tested variants share only the pre-attention normalization, only the spatial-reduction convolution, only the post-spatial-reduction LayerNorm, the complete spatial-reduction path, or the complete recipient preprocessing path.

This experiment is used for mechanistic identification, not for choosing the final hub count. Its outcome defines an architectural constraint for all subsequent hub models: relocated attention retains its donor-specific pre-normalization, spatial-reduction operation, and post-reduction normalization rather than being routed through a fully shared recipient path.

\subsection{Hub Frontier and Frozen Operating Points}
All attention-hub models start from the same frozen State120 parent and retain the same total budget of 120 Stage-3 attention heads. We construct a nested execution-depth frontier
\[
K\in\{20,16,14,12,10\}.
\]
Relocation is constrained to a maximum displacement of four original Stage-3 depths. Hub layouts are constructed as a nested parent-to-child trajectory. Donor-specific preprocessing identified by the mechanism experiment is preserved in all hub layouts.

Before the final replicated evaluation, K20, K14, and K10 are assigned fixed roles as conservative, balanced, and aggressive operating points, respectively. These roles are frozen before the final three-seed recovery comparison.

\subsection{Final Recovery Protocol}
The final K20, K14, and K10 configurations are evaluated using fixed seeds $\{42,43,44\}$ and a fixed recovery duration of three epochs. Recovery uses the 4,625-image fitting subset. Parameters of relocated attention heads are optimized with learning rate $1\times10^{-5}$, while the corresponding pre-hub FFN components use learning rate $2\times10^{-6}$. Only parameters directly involved in relocation and local representation adaptation are updated; the remainder of the network is fixed.

The recovery endpoint is predetermined at epoch 3 for every model and seed. We do not select the best epoch or best seed based on external-development performance. Final results are reported as the mean and standard deviation over the three fixed runs.

\subsection{Layout-Matched State120 Control}
For each final hub layout, we construct a control that retains the original State120 architecture: 120 heads at 40 attention depths. No head is relocated and no attention execution point is removed. However, the control trains the same classes of components adapted in the corresponding hub model: the Q/K/V/O parameters of retained donor heads that would be relocated and the same immediately preceding FFN components used by hub recovery.

The matched control uses the same fitting images, seeds, fixed three-epoch duration, optimizer policy, learning-rate policy, and evaluation endpoint as the corresponding hub configuration. The primary architectural comparison is therefore State120 with 120 heads at 40 attention depths versus a hub model with the same 120 heads at $K$ attention depths.

\subsection{Quality Metrics and Uncertainty}
Segmentation quality is evaluated using mean Intersection over Union (mIoU), person-class IoU, background-class IoU, and pixel accuracy. The primary metric is mIoU.

For the final architectural comparisons, we estimate image-sampling uncertainty using a paired image-level bootstrap with 20,000 resamples. The same resampled image indices are applied to the paired hub and State120 predictions. Because the K20 confidence interval includes zero, its positive point estimate is not interpreted as statistical superiority. Instead, it supports the conservative conclusion that consolidating the same 120-head budget from 40 to 20 execution depths incurs no measurable quality penalty under the matched protocol.

The bootstrap quantifies image-sampling uncertainty conditional on the fixed trained models and the current external development observer; it does not estimate the full distribution over arbitrary training seeds or repeated architecture searches.

\subsection{Parameter and Compute Accounting}
We report total model parameters, retained Stage-3 attention heads, Stage-3 attention execution depths, multiply-accumulate operations (MACs), and the corresponding FLOP estimate. We use the convention that one MAC is one multiply-accumulate pair and approximately two FLOPs.

The MAC count includes linear layers, Conv2d layers, attention score multiplication $QK^{\top}$, and attention--value multiplication $AV$. We exclude bias additions, LayerNorm operations, softmax, elementwise nonlinearities, reshaping and transposition operations, and final evaluation interpolation.

Under this convention, State120 and the K20/K14/K10 hub configurations have essentially identical retained-head arithmetic: State120 and K20 both require approximately 102.769 GMAC. Thus, the State120-to-hub transformation is primarily a change in execution topology rather than additional arithmetic pruning.

\subsection{Deployment-Equivalence Evaluation}
FOLDED\_VEC is an inference-only deployment transformation and therefore receives a separate functional-equivalence evaluation. We compare the physical K20 implementation with its FOLDED\_VEC realization using all 572 external-development images, all three fixed K20 recovery seeds, FP16 inference, no additional training, no architecture search, and no parameter selection.

We report per-seed mIoU, folded-minus-physical mIoU difference, raw-logit RMSE, maximum absolute logit difference, the number of pixels whose argmax prediction changes, and the fraction of changed argmax predictions. Because floating-point operation ordering changes, we do not require bitwise equality; the experiment determines whether FOLDED\_VEC preserves the validated K20 function to measurement precision.

\subsection{Hardware Benchmark}
Final latency experiments are performed on an NVIDIA L4 using $512\times512$ inputs. All compared implementations use FP16 autocast, channels-last memory layout, \texttt{torch.compile(mode="max-autotune")}, synchronized CUDA timing, and batch sizes 1, 4, and 8.

For every model--batch-size pair, the TorchDynamo/compiler state is reset before compilation. This prevents specialization-cache and recompilation-limit effects from contaminating cross-model comparisons. A dedicated diagnostic showed that an earlier apparent K14/K10 slowdown was a TorchDynamo recompilation-limit artifact rather than an architectural property.

The final hardware analysis follows the same fixed-capacity hierarchy as the functional evaluation: State120 $\rightarrow$ physical K20 isolates the effect of logical depth consolidation, physical K20 $\rightarrow$ K20 FOLDED isolates the deployment reparameterization, and State120 $\rightarrow$ physical K20 $\rightarrow$ K20 FOLDED reports the complete deployment path.

\subsection{Evaluation Hierarchy}
The experimental design separates two questions. First, \emph{functional consolidation} is evaluated by comparing each hub configuration with its layout-matched State120 control. Both models retain the same 120 Stage-3 attention heads and receive matched recovery. This comparison isolates the effect of reorganizing retained attention from the original 40 execution depths into fewer attention hubs.

Second, \emph{deployment realization} is evaluated by comparing the validated physical K20 implementation with its FOLDED\_VEC realization. This comparison keeps the hub architecture fixed and isolates the effect of algebraic folding and vectorized execution. Accordingly, the primary evaluation chain is
\[
\text{State120}\rightarrow\text{K20}\rightarrow\text{K20 FOLDED},
\]
representing fixed-capacity reference, depth-to-width consolidation, and hardware realization.

\section{Results}
\subsection{Head Thinning Produces a Fixed 120-Head Parent}
We first reduce Stage-3 attention capacity before studying execution-depth redundancy. The source Stage-3 attention population contains 200 heads. Sequential thinning initially improves the observed external-development mIoU, reaching its highest point near 150 heads, after which the trajectory becomes non-monotonic. At 120 heads, the resulting model reaches mIoU 0.959512.

\begin{table}[t]
\centering
\caption{Stage-3 head-thinning trajectory on the external development observer.}
\label{tab:head_thinning}
\begin{tabular}{rr}
\toprule
Retained heads & External mIoU \\
\midrule
200 & 0.960035 \\
190 & 0.960246 \\
180 & 0.960513 \\
170 & 0.960523 \\
160 & 0.960701 \\
150 & 0.960824 \\
140 & 0.960128 \\
130 & 0.959863 \\
120 & 0.959512 \\
100 & 0.959884 \\
80  & 0.958381 \\
\bottomrule
\end{tabular}
\end{table}

For all subsequent experiments, we freeze the predeclared 120-head state, denoted State120. This point is not selected because it maximizes the external observer; its purpose is to provide a fixed retained-attention budget for studying whether the same 120 functions require all of their original execution depths.

\subsection{Some Attention Functions Are Locally Mobile Across Depth}
The mobility study covers all 200 original Stage-3 heads and 390 directed adjacent relocations. For a subset of task-relevant heads, moving the pretrained function to a nearby depth preserves useful behavior that would otherwise be lost by dropping the same head.

Across all relocations, MOVE outperforms DROP in 181 of 390 directed cases. The relationship becomes much stronger when conditioning on source-head importance: for DROP damage at least $2\times10^{-5}$, MOVE beats DROP in 68/108 cases (63.0\%); at least $5\times10^{-5}$, 38/42 (90.5\%); at least $10^{-4}$, 20/22 (90.9\%); and at least $2\times10^{-4}$, 6/6. The Spearman correlation between DROP damage and MOVE gain over DROP is approximately $\rho=0.459$.

The corrected SegFormer-native reinitialization control prevents a stronger interpretation: pretrained MOVE is not universally superior to inserting a newly initialized head. We therefore restrict the claim to the existence of local functional depth mobility for some task-relevant attention heads.

\subsection{Successful Relocation Requires More Than Head Parameters}
We evaluate relocation adaptation on 24 heads under RAW MOVE, HEAD\_ONLY\_FT, and HEAD+PRE\_FFN\_FT. The mean calibration-mIoU changes relative to the parent are $-0.000025$, $-0.000067$, and $+0.000141$, respectively. Joint head-plus-boundary adaptation is above the parent in 24/24 cases and is better than HEAD\_ONLY\_FT in 24/24 cases.

\begin{table}[t]
\centering
\caption{Relocation adaptation across 24 heads.}
\label{tab:reloc_adaptation}
\begin{tabular}{lrrr}
\toprule
Variant & Mean $\Delta$ mIoU & Median $\Delta$ mIoU & Better than parent \\
\midrule
RAW MOVE & -0.000025 & -0.000017 & 0/24 \\
HEAD\_ONLY\_FT & -0.000067 & -0.000044 & 1/24 \\
HEAD+PRE\_FFN\_FT & +0.000141 & +0.000130 & 24/24 \\
\bottomrule
\end{tabular}
\end{table}

The result does not identify the preceding FFN as a uniquely sufficient mechanism. It shows that relocation is not well described as a head-only operation: the representation boundary around the relocated function matters.

\subsection{Mobility Is Conditioned on the Donor Preprocessing Path}
With the exact donor-specific path preserved, the mechanism configuration reaches external mIoU 0.960737. Replacing components of this path with shared or recipient-side alternatives consistently degrades performance.

\begin{table}[t]
\centering
\caption{Donor-path mechanism ablation. Values are external mIoU changes relative to exact donor-path execution.}
\label{tab:donor_path}
\begin{tabular}{lr}
\toprule
Preprocessing configuration & $\Delta$ mIoU \\
\midrule
Exact donor path & 0 \\
Shared pre-norm only & -0.001853 \\
Shared SR convolution only & -0.020608 \\
Shared post-SR LayerNorm only & -0.012518 \\
Shared full SR path & -0.022242 \\
Recipient preprocessing path & -0.022711 \\
\bottomrule
\end{tabular}
\end{table}

The largest losses occur when the spatial-reduction convolution or complete recipient preprocessing path is shared. These results reveal an important asymmetry: an attention function may be locally mobile across depth, while the preprocessing path supporting that function remains depth-specific. We therefore describe the observed phenomenon as path-conditioned functional attention mobility.

\begin{figure*}[t]
\centering
\includegraphics[width=\textwidth]{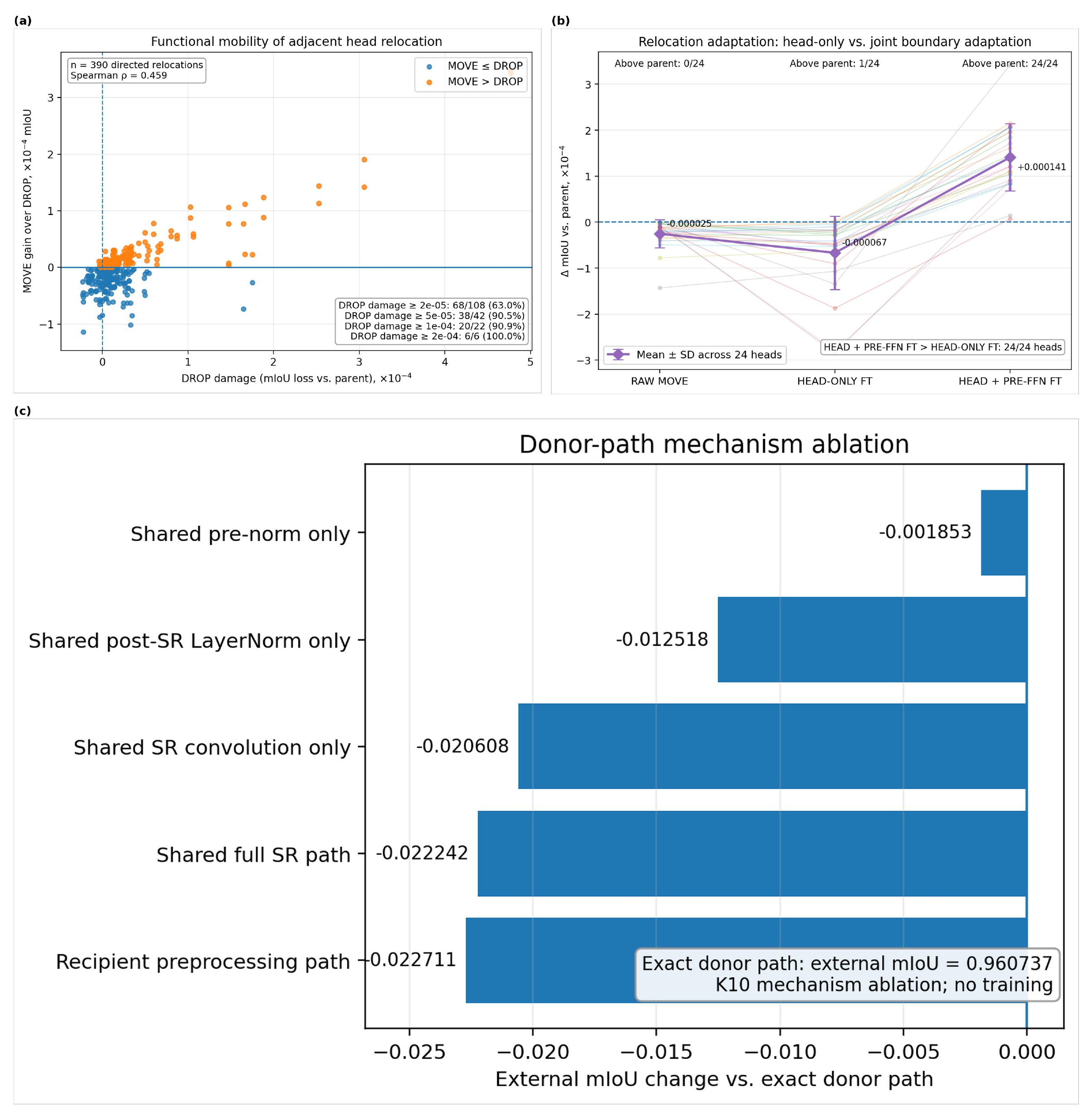}
\caption{Evidence for path-conditioned functional attention mobility. (a) Functional mobility of directed adjacent head relocation. (b) Relocation adaptation across 24 heads. (c) Donor-path mechanism ablation.}
\label{fig:mobility_mechanism}
\end{figure*}

\subsection{The Same 120 Heads Can Be Consolidated into Fewer Execution Depths}
All hub configurations contain exactly the same 120 retained heads as State120. Only their distribution across execution depths changes.

\begin{table}[t]
\centering
\caption{Fixed-capacity execution-hub frontier. Every configuration retains the same 120 Stage-3 heads.}
\label{tab:hub_frontier}
\begin{tabular}{lrrr}
\toprule
Configuration & Retained heads & Attention depths & External mIoU \\
\midrule
State120 & 120 & 40 & 0.959512 \\
K20 & 120 & 20 & 0.960039 \\
K16 & 120 & 16 & 0.959692 \\
K14 & 120 & 14 & 0.959753 \\
K12 & 120 & 12 & 0.958146 \\
K10 & 120 & 10 & 0.958877 \\
\bottomrule
\end{tabular}
\end{table}

The key result is that attention execution depth can initially be reduced substantially without removing any additional attention function. K20 halves the number of Stage-3 attention execution depths, $40\rightarrow20$, while retaining all 120 heads. More aggressive consolidation eventually degrades quality, indicating that attention capacity and attention execution depth are separable but not completely independent.

\begin{figure*}[t]
\centering
\includegraphics[width=\textwidth]{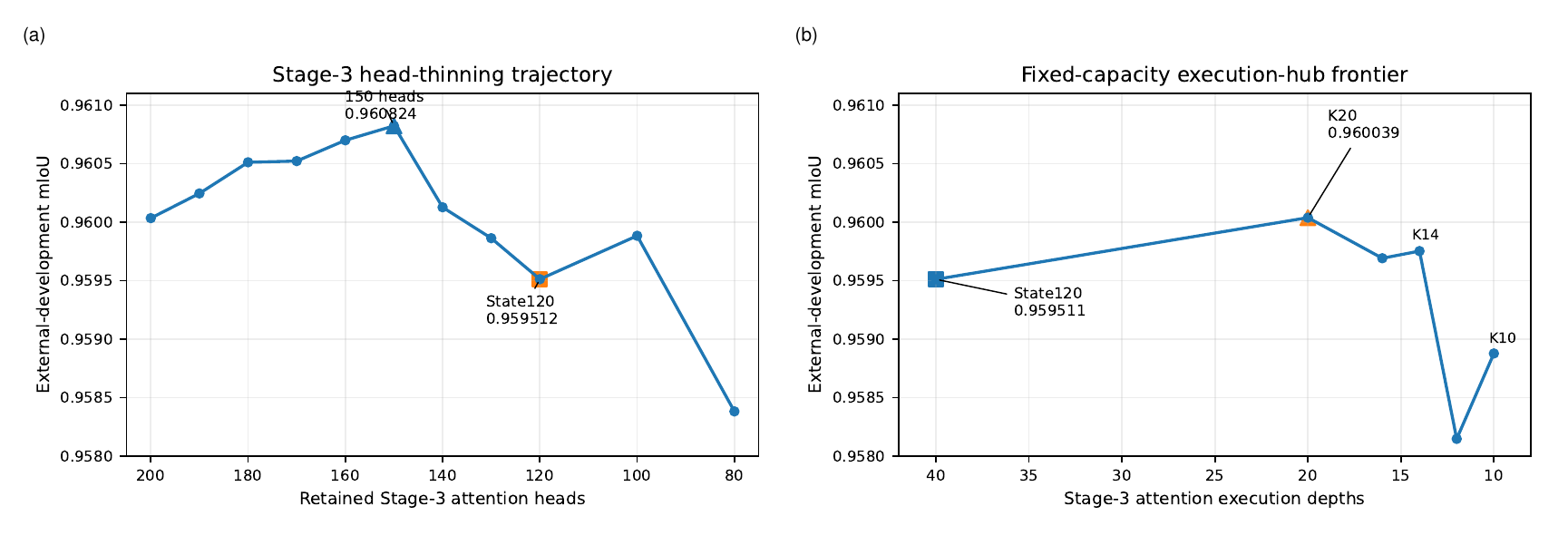}
\caption{Attention-capacity thinning and fixed-capacity execution-depth consolidation. (a) External-development mIoU along the sequential Stage-3 head-thinning trajectory. (b) Execution-hub frontier with retained attention capacity fixed at the same 120 heads.}
\label{fig:thinning_hubs}
\end{figure*}

\subsection{K20 Preserves Quality Under the Layout-Matched State120 Control}
The main causal comparison uses layout-matched State120 controls. For each hub layout, the corresponding control retains the same 120 heads at the original 40 attention depths while receiving the same fixed recovery opportunity and matched trainable component classes.

\begin{table}[t]
\centering
\caption{Three-seed recovered hub models versus layout-matched State120 controls.}
\label{tab:matched_control}
\begin{tabular}{lrrr}
\toprule
Configuration & Matched State120 & Hub mIoU & Paired $\Delta$ \\
\midrule
K20 & 0.960400 & 0.960686 & +0.000285 \\
K14 & 0.960422 & 0.960169 & -0.000253 \\
K10 & 0.960395 & 0.959132 & -0.001263 \\
\bottomrule
\end{tabular}
\end{table}

K20 is higher than its layout-matched State120 control in all three fixed-seed runs. Its recovered external mIoU is $0.960686\pm0.000059$, compared with $0.960400\pm0.000074$ for the matched State120 control. This central comparison holds retained attention capacity fixed: 120 heads at 40 depths versus 120 heads at 20 depths.

\subsection{Bootstrap Analysis Supports a No-Measurable-Penalty Interpretation}
The positive K20 point estimate should not be interpreted as evidence that consolidation improves segmentation quality. Across 20,000 paired image-bootstrap resamples, K20 versus matched State120 has bootstrap mean $+0.000283$ and 95\% confidence interval $[-0.000481,+0.001000]$. The interval includes zero.

For K14 and K10, the corresponding intervals are $[-0.002333,+0.001313]$ and $[-0.005565,+0.002409]$. Accordingly, our main interpretation is that consolidating the same 120-head attention budget from 40 to 20 execution depths incurs no measurable quality penalty under the matched recovery protocol. K20's slightly higher point estimate is reported descriptively, not as statistical superiority.

\begin{figure}[t]
\centering
\includegraphics[width=\linewidth]{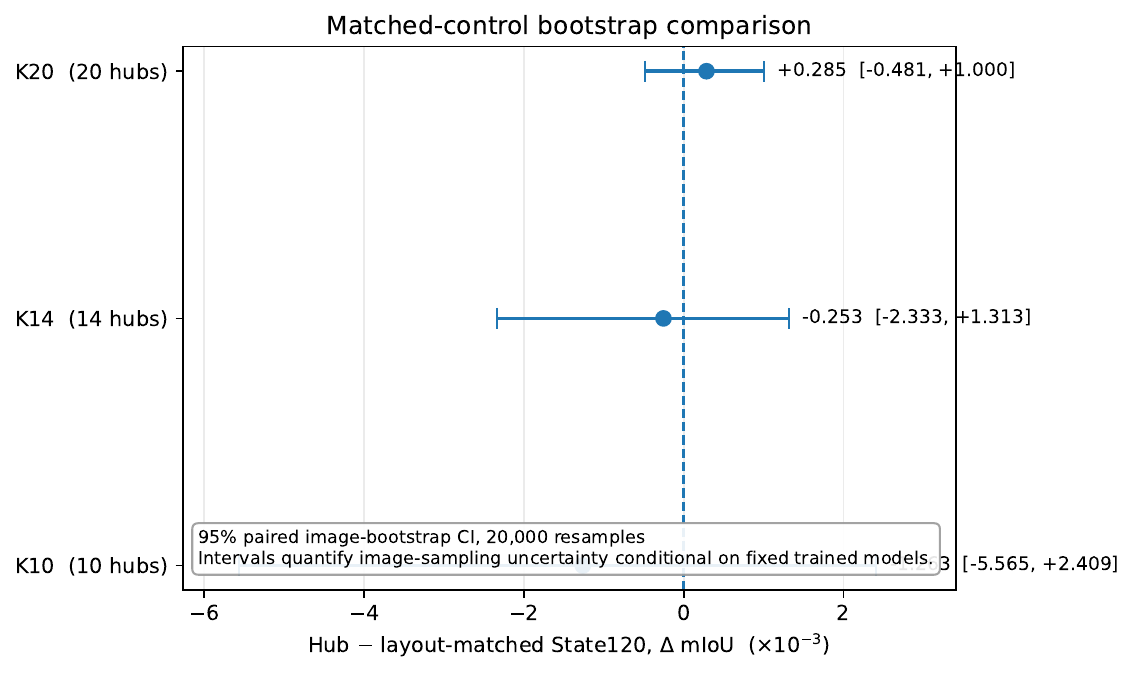}
\caption{Matched-control comparison of attention-hub configurations. Points show paired mIoU differences versus layout-matched State120 controls; horizontal bars denote 95\% paired image-bootstrap confidence intervals.}
\label{fig:matched_control}
\end{figure}

\subsection{More Aggressive Consolidation Crosses a Quality Frontier}
The matched controls reveal a trend beyond K20. K14 falls below its matched State120 control by $-0.000253$, while K10 falls below its matched control by $-0.001263$ and also exhibits larger seed variability. Thus, the useful regime is not ``as few hubs as possible.'' K20 is the strongest quality-preserving operating point in the tested frontier, while further consolidation increasingly trades quality for reduced sequential depth.

\subsection{Hub Consolidation Changes Execution Depth, Not Retained Arithmetic}
State120 contains 120 heads executed across 40 attention depths. K20 contains essentially the same parameters and the same retained-head arithmetic, but executes those 120 heads across only 20 attention depths.

\begin{table}[t]
\centering
\caption{Parameter and compute accounting.}
\label{tab:complexity}
\begin{tabular}{lrrrr}
\toprule
Model & Params & Heads & Attention depths & GMAC \\
\midrule
State120 & 78.026M & 120 & 40 & 102.769 \\
K20 & 78.020M & 120 & 20 & 102.769 \\
K14 & 78.018M & 120 & 14 & 102.769 \\
K10 & 78.016M & 120 & 10 & 102.769 \\
\bottomrule
\end{tabular}
\end{table}

Therefore, State120 $\rightarrow$ K20 is principally a transformation of execution topology, not a reduction in retained arithmetic. This distinction is central: the matched quality result cannot be explained by using fewer heads or substantially fewer MACs.

\subsection{Logical Depth Reduction Alone Produces Only a Small Hardware Gain}
The physical hub implementation initially preserves donor-specific preprocessing literally. Under clean compilation, State120 and physical K20 have very similar whole-model latency.

\begin{table}[t]
\centering
\caption{Same-run NVIDIA L4 latency for State120 and the physical K20 implementation.}
\label{tab:physical_latency}
\begin{tabular}{rrr}
\toprule
Batch & State120 (ms) & Physical K20 (ms) \\
\midrule
1 & 17.925 & 17.711 \\
4 & 50.260 & 50.085 \\
8 & 99.448 & 99.261 \\
\bottomrule
\end{tabular}
\end{table}

This corresponds to reductions of approximately 1.19\%, 0.35\%, and 0.19\% at batch sizes 1, 4, and 8. Thus, halving logical attention depth does not automatically halve latency. All 120 attention functions still need to be evaluated, and donor-specific preprocessing remains fragmented.

\subsection{FOLDED\_VEC Converts the Structural Reduction into a Measurable Deployment Gain}
FOLDED\_VEC reorganizes the physical execution of donor-specific paths without replacing them by shared approximations.

\begin{table}[t]
\centering
\caption{Same-run NVIDIA L4 latency through the deployment pipeline.}
\label{tab:folded_latency}
\begin{tabular}{rrrr}
\toprule
Batch & State120 (ms) & Physical K20 (ms) & K20 FOLDED (ms) \\
\midrule
1 & 17.925 & 17.711 & 16.979 \\
4 & 50.260 & 50.085 & 48.679 \\
8 & 99.448 & 99.261 & 95.888 \\
\bottomrule
\end{tabular}
\end{table}

Relative to the same 120-head State120 parent, the final folded implementation reduces end-to-end latency by approximately 5.28\%, 3.15\%, and 3.58\% at batch sizes 1, 4, and 8, respectively.

\begin{figure}[t]
\centering
\includegraphics[width=\linewidth]{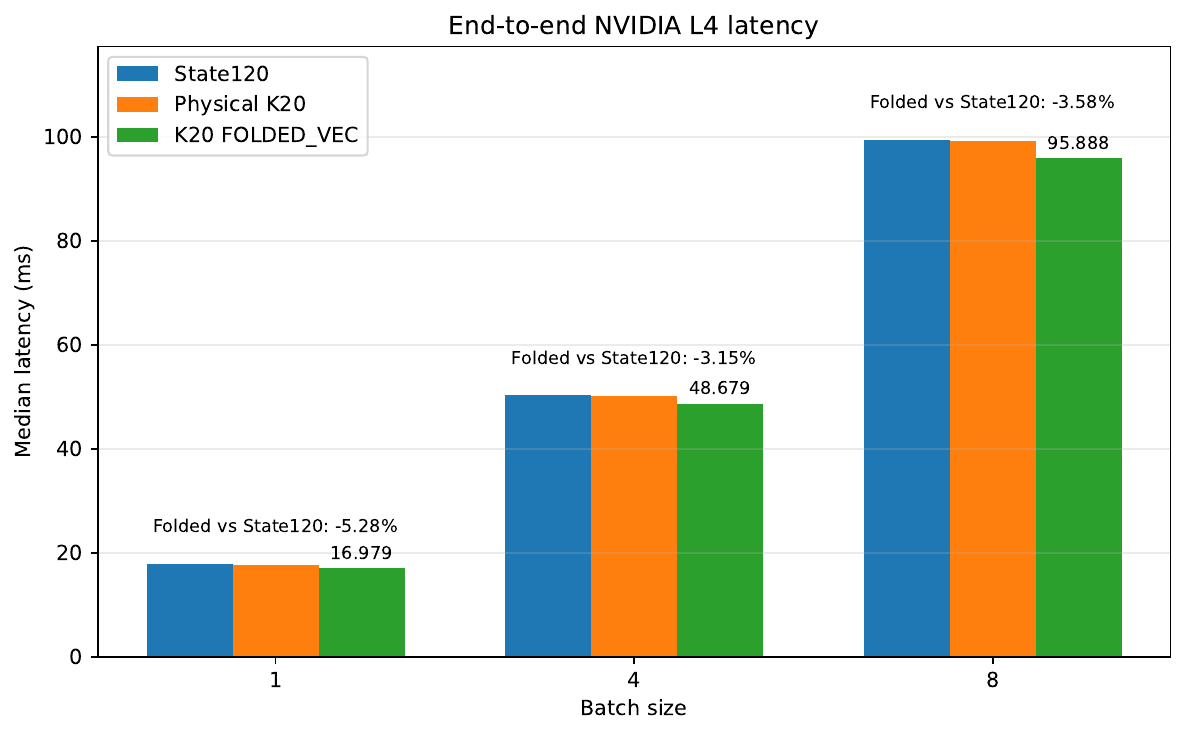}
\caption{End-to-end NVIDIA L4 latency for State120, physical K20, and K20 FOLDED\_VEC at batch sizes 1, 4, and 8.}
\label{fig:latency}
\end{figure}

\subsection{FOLDED\_VEC Preserves K20 Quality to Measurement Precision}
Because FOLDED\_VEC changes physical execution order, we verify its output rather than assume exact equivalence. Across all 572 external-development images and all three fixed K20 seeds,
\[
\mathrm{mIoU}_{\mathrm{physical}}=0.960685013,\qquad
\mathrm{mIoU}_{\mathrm{folded}}=0.960684432.
\]
The mean change is $-5.81\times10^{-7}$. Only 2,582 of 449,839,104 valid pixel predictions change their argmax label, corresponding to 0.000574\%. Mean raw-logit RMSE is 0.001081, and the maximum absolute seed-level mIoU difference is $2.02\times10^{-6}$. We therefore treat FOLDED\_VEC as quality-preserving to measurement precision rather than bitwise identical.

\subsection{The Depth-to-Width Frontier Saturates Beyond K20}
K14 and K10 reduce attention execution depth further, but their whole-model latency remains close to K20 while both lose quality relative to their matched State120 controls. As the hub count decreases, each hub becomes wider and must execute more donor-specific preprocessing and projection work. Reduced sequential depth is increasingly offset by wider parallel operations.

Consequently, K20 is the strongest operating point in the tested regime: it retains the same 120-head budget as State120, reduces attention execution depths from 40 to 20, shows no measurable quality penalty under the matched control, behaves stably across the three fixed seeds, and yields approximately 3.1--5.3\% lower end-to-end latency after deployment folding.

\section{Discussion}
\subsection{Attention Redundancy Has Both a Capacity and an Execution-Depth Dimension}
The results suggest that attention redundancy in a hierarchical vision transformer is not adequately described by head count alone. Head thinning identifies capacity redundancy: some learned attention functions can be removed without materially harming the downstream task. The State120 parent already exploits this form of redundancy by retaining only 120 Stage-3 heads.

The hub experiments reveal a second form of redundancy. Even after retained attention capacity is fixed, the same functions do not necessarily require all of their original sequential execution locations. State120 executes 120 retained heads across 40 attention depths, whereas K20 executes the same 120-head budget across 20 attention hubs. Under the layout-matched recovery comparison, this two-fold reduction in attention execution depth produces no measurable quality penalty.

This motivates a distinction between \emph{attention capacity redundancy} and \emph{attention execution-depth redundancy}. Removing additional heads changes the set of retained functions; hub consolidation changes where those retained functions are executed. The distinction is reflected in analytical compute: State120 and K20 have essentially the same retained-head MAC count even though K20 uses half as many Stage-3 attention execution depths.

\subsection{Functional Mobility Does Not Imply Layer Interchangeability}
Some attention heads preserve useful task function when moved to nearby depths, showing that their function is not rigidly attached to a single original location. However, neither the relocation adaptation experiments nor the donor-path ablations support treating neighboring attention layers as interchangeable.

Head-only fine-tuning is insufficient in the controlled relocation study, while joint adaptation of the relocated head and its preceding representation boundary is consistently more effective. The donor-path mechanism experiment provides even stronger evidence: replacing donor-specific spatial-reduction components with shared or recipient-side preprocessing causes losses exceeding 0.02 mIoU in some tested configurations.

These findings suggest that the effective function of an attention head is not fully represented by its Q/K/V/O parameters alone. A more appropriate conceptual unit is the composition of representation state, donor preprocessing, attention, and the surrounding representation boundary. This reconciles two observations: useful attention can move across depth, while naively sharing or replacing its depth-specific path can fail badly.

\subsection{Why K20 Is a Natural Operating Point}
The nested hub frontier shows that decreasing the number of attention execution depths is not monotonically beneficial. K20 reduces the original 40 attention execution depths to 20 while preserving quality under the matched comparison. More aggressive K14 and K10 consolidation moves below the corresponding State120 controls.

This suggests that functional mobility has a practical radius and capacity limit. A retained head can tolerate some change in execution context, but repeatedly moving more heads and concentrating more heterogeneous functions into fewer hubs eventually changes the representation trajectory too strongly for short local recovery to compensate. The result therefore does not imply that attention depth is simply ``50\% redundant''; it supports the narrower claim that the tested 120 retained functions can be reorganized from 40 to 20 execution depths without measurable degradation under the matched protocol.

\subsection{Functional Compression and Hardware Compression Are Distinct}
K20 halves the number of logical Stage-3 attention execution depths relative to State120, yet the initial physical implementation provides only a small end-to-end latency reduction. The same 120 retained attention functions are still computed, and donor-specific preprocessing required for correctness remains heterogeneous.

This creates a tension between functional preservation and regular hardware execution. FOLDED\_VEC addresses the tension without replacing heterogeneous donor paths by a shared approximation. It algebraically folds and vectorizes compatible operations while preserving donor-specific parameters. The resulting end-to-end reduction relative to State120 is 5.28\%, 3.15\%, and 3.58\% at batch sizes 1, 4, and 8, while the full external-development equivalence evaluation shows only measurement-level output differences.

\subsection{Fewer Sequential Attention Points Need Not Mean Fewer FLOPs}
State120 and K20 both contain 120 retained heads and both require approximately 102.769 GMAC under our counting convention, despite having 40 and 20 attention execution depths. The reduction in sequential execution depth changes the dependency structure of the computation rather than eliminating the underlying retained-head arithmetic.

This explains why the latency improvement is much smaller than the two-fold reduction in logical depth. Wider hubs also concentrate preprocessing and projection work. K14 and K10 reinforce this point: beyond K20, reducing the hub count further yields little stable end-to-end latency benefit while quality declines. The hardware frontier therefore exhibits a depth-to-width saturation point.

\subsection{Relation to Cross-Layer Sharing and Transfer}
The results provide a middle ground between treating layers as distinct sequential transformations and exploiting cross-layer similarity through shared or approximated attention. The useful pretrained attention function itself may be locally relocatable, but the computational context associated with that function remains partially depth-specific. This is why preserving donor identity is important even when several donor functions are executed at the same hub.

Attention hubs therefore differ from replacing several depths with one shared attention function. A hub is a common execution location for heterogeneous retained functions. Similarity between attention maps need not imply that complete attention modules can be substituted for one another when normalization, spatial reduction, and representation state differ.

\subsection{Implications for Transformer Compression}
A conventional compression pipeline asks which heads and which layers should be deleted. Our experiments suggest an intermediate question: which useful functions can be relocated before their original execution depth is removed? This produces a broader structural action space: remove, retain in place, or relocate.

Such a formulation can distinguish between a transformation that is unnecessary and one that remains useful but does not require its original sequential location. The present work does not establish that this formulation generalizes to all Transformers, but it provides evidence that deletion and relocation are meaningfully different operations in the tested SegFormer setting.

\subsection{What the Positive K20 Point Estimate Does and Does Not Mean}
K20 has a slightly higher external-development mIoU than its matched State120 control in all three fixed seeds. The observed paired point difference is approximately $+0.000285$. It would be tempting to interpret this as evidence that consolidation itself improves generalization.

The uncertainty analysis does not support that conclusion. The paired image-bootstrap confidence interval includes zero. The appropriate interpretation is therefore that K20 does not show a measurable quality penalty under the matched protocol. We report the positive point estimate descriptively but do not claim statistical superiority.

\section{Limitations}

\subsection{Single Downstream Task and Dataset}

The experiments use one binary human-segmentation task on the Supervisely Persons Dataset.

This setting is useful because the deep SegFormer-B5 encoder provides substantial structural redundancy to study, but the observed mobility and consolidation frontier may depend on the task. Binary human segmentation has a substantially smaller output space than datasets such as ADE20K or Cityscapes.

We therefore do not claim that the same heads, hub locations, or compression frontier would transfer unchanged to other segmentation tasks.

Validation on additional dense-prediction datasets is the most important next test of external validity.

\subsection{Single Source Architecture}

The study focuses on SegFormer-B5.

Although the mechanism is motivated at the level of hierarchical Transformers, SegFormer's spatial-reduction attention is architecturally specific. The importance of preserving donor-specific SR paths may not translate directly to architectures using standard full-resolution self-attention, windowed attention, or different hierarchical mechanisms.

Accordingly, path-conditioned mobility should currently be interpreted as an empirical property demonstrated in the tested SegFormer setting, rather than a universal property of Transformer attention.

Testing additional SegFormer scales and non-SegFormer vision transformers would be required to establish broader architectural generality.

\subsection{Stage-3 Focus}

All structural attention modifications are concentrated in Stage 3.

This stage is particularly suitable for the study because it contains 40 Transformer blocks in B5 and therefore exposes a large amount of repeated attention depth.

The experiments do not establish that attention in earlier or later SegFormer stages has the same mobility properties.

Cross-stage relocation is also not evaluated.

Thus, our claim concerns local within-stage mobility, not arbitrary relocation throughout the full encoder.

\subsection{Mobility Is Tested Over a Restricted Distance}

The hub construction uses bounded local relocation, with a maximum movement radius of four original depths.

The broader mobility analysis also emphasizes adjacent and nearby transfers.

The results therefore do not support the claim that attention functions are freely portable over long distances. In fact, the degradation observed under stronger hub consolidation is consistent with the opposite interpretation: mobility is local and becomes less reliable as representation context changes.

\subsection{External Development Observer Is Not an Untouched Test Set}

The 572-image external split has been repeatedly observed during the broader research process.

Although final architecture roles, recovery endpoints, and fixed seeds were not selected from this split, it cannot be considered a fully sequestered test set.

We therefore consistently report it as an external development observer rather than an unseen test set.

A stronger final evaluation would use an independently sequestered test set or an entirely separate segmentation dataset that was never inspected during method development.

This is one of the most important limitations of the present evidence.

\subsection{Limited Number of Recovery Seeds}

The final recovery comparison uses three fixed seeds.

The low variance of K20 across these runs is encouraging, but three seeds are insufficient to characterize the full distribution of optimization outcomes.

The paired image bootstrap quantifies uncertainty arising from image sampling conditional on the already trained models. It does not estimate uncertainty over arbitrary training seeds or repeated architectural searches.

For this reason, we use the bootstrap to support a no-measurable-penalty interpretation rather than a claim of statistical superiority.

\subsection{The Head-Thinning Trajectory Is Path Dependent}

The long head-thinning trajectory is explicitly non-monotonic.

The externally strongest point occurs near 150 heads, while later states can partially recover after additional removals.

This means that no universal ``safe head count'' can be inferred from the trajectory.

State120 is useful as a fixed experimental parent, but the study does not establish that 120 is an optimal retained-head budget for SegFormer-B5 generally.

A different selection rule, training history, downstream task, or random seed could produce a different retained population.

\subsection{Hardware Results Depend on the Execution Stack}

Latency is measured on a single NVIDIA L4 software/hardware stack using FP16, channels-last tensors, and compiled PyTorch execution.

The benchmark required explicit control of compiler specialization state; an earlier apparent K14/K10 slowdown was shown to be a TorchDynamo recompilation-limit artifact rather than a model property.

This illustrates that reported hardware gains are sensitive to compiler and runtime behavior.

The approximately 3--5\% latency reduction relative to State120 should therefore not be assumed to transfer quantitatively to other GPUs, inference engines, batch sizes, or compiler versions.

Evaluation on additional hardware would strengthen the systems conclusion.

\subsection{More Aggressive Folded Models Were Not Fully Revalidated}

K20 receives a full FOLDED\_VEC external-development equivalence evaluation across all three fixed seeds.

K14 and K10 pass numerical sanity checks for their folded graphs, but they do not receive the same full 572-image, three-seed folded validation because they are not selected as final deployment configurations.

Their hardware results are therefore used only to diagnose depth-to-width saturation, not as fully validated deployment operating points.

\subsection{No State-of-the-Art Compression Claim}

The goal of this study is not to establish the strongest available SegFormer compression ratio or GPU speedup.

Methods operating along other dimensions---such as token merging, structured head pruning, dynamic token reduction, quantization, or architecture-specific kernel design---may achieve larger absolute throughput gains.

Our experiments instead isolate a narrower scientific question: whether retained attention capacity must remain distributed over its original sequential depth.

Consequently, the reported latency improvement should be interpreted as evidence that the proposed structural transformation can be realized efficiently, rather than as a claim of state-of-the-art inference acceleration.

\subsection{Donor-Path Preservation Limits Immediate Hardware Savings}

The same mechanism that enables functional relocation also constrains deployment efficiency.

Preserving donor-specific preprocessing introduces heterogeneous operations that are harder to fuse than a fully shared recipient path. The latter would be computationally simpler but performs substantially worse in the mechanism experiment.

FOLDED\_VEC reduces this overhead, but the remaining latency gain is much smaller than the reduction in logical attention depth.

This trade-off is fundamental to the current implementation: preserving functional identity and maximizing execution regularity are partially competing objectives.

Future implementations may achieve better kernels for heterogeneous donor groups without changing the learned model.

\subsection{Broader Generalization Remains the Main Open Question}

The present experiments support the following case-specific conclusion:

a fixed 120-head Stage-3 attention budget in the tested SegFormer-B5 human-segmentation model can be reorganized from 40 to 20 attention execution depths without measurable quality degradation under the matched recovery protocol.

The strongest next step is not a more aggressive hub configuration on the same model. It is to determine whether the same phenomenon appears across:

\begin{itemize}
  \item additional segmentation datasets,
  \item other SegFormer scales,
  \item other hierarchical vision transformers,
  \item and potentially Transformer families outside computer vision.
\end{itemize}
Such experiments would determine whether path-conditioned functional mobility is a SegFormer-specific phenomenon or a broader property of learned Transformer representations.

\section{Conclusion}
We studied whether retained attention capacity in a hierarchical vision transformer must remain distributed across the sequential depths at which it was originally learned. Using SegFormer-B5 Stage 3 as a controlled setting, we find that some task-relevant attention functions exhibit local functional mobility across depth, but this mobility is not equivalent to unrestricted layer interchangeability. Successful relocation depends on the surrounding representation context, and preserving donor-specific normalization and spatial-reduction preprocessing is particularly important.

We exploit this path-conditioned mobility to separate attention capacity from attention execution depth. Starting from a fixed State120 parent with 120 retained heads distributed across 40 attention depths, we consolidate the same head budget into 20 execution hubs. Under a layout-matched recovery control, K20 shows no measurable quality penalty: the paired point difference is $+0.000285$, while the paired image-bootstrap confidence interval includes zero. More aggressive consolidation progressively loses quality, indicating that functional mobility is structured rather than unlimited.

The transformation primarily changes execution topology rather than retained arithmetic. State120 and K20 contain the same 120-head budget and essentially the same analytical MAC count, while K20 halves the number of Stage-3 attention execution depths. A deployment-only folded vectorized realization preserves the validated K20 behavior to measurement precision and converts this structural reorganization into an approximately 3.1--5.3\% end-to-end latency reduction relative to the same State120 parent in our NVIDIA L4 measurements.

These results suggest that Transformer compression need not be framed only in terms of which functions can be deleted. A complementary question is which useful functions can be relocated before their original execution depth is removed. In the tested SegFormer setting, this exposes a form of path-conditioned execution-depth redundancy distinct from conventional head pruning and cross-layer attention sharing. Establishing how broadly this phenomenon extends across tasks and architectures remains an important direction for future work.

\clearpage
\appendix
\section{Supplementary Figures}
\begin{figure}[ht]
\centering
\includegraphics[width=0.92\linewidth]{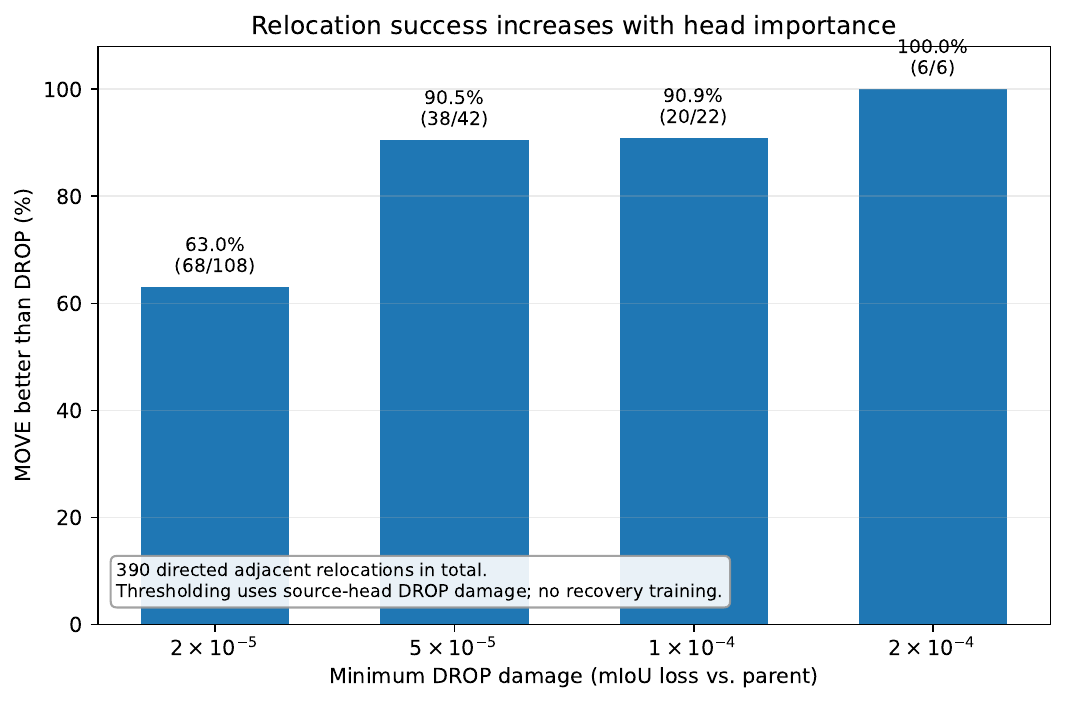}
\caption{Relocation success conditioned on source-head importance. Bars show the fraction of directed adjacent relocations for which MOVE outperforms DROP after restricting the analysis to heads whose removal causes at least the indicated mIoU loss.}
\label{fig:supp_mobility_threshold}
\end{figure}

\begin{figure}[ht]
\centering
\includegraphics[width=0.92\linewidth]{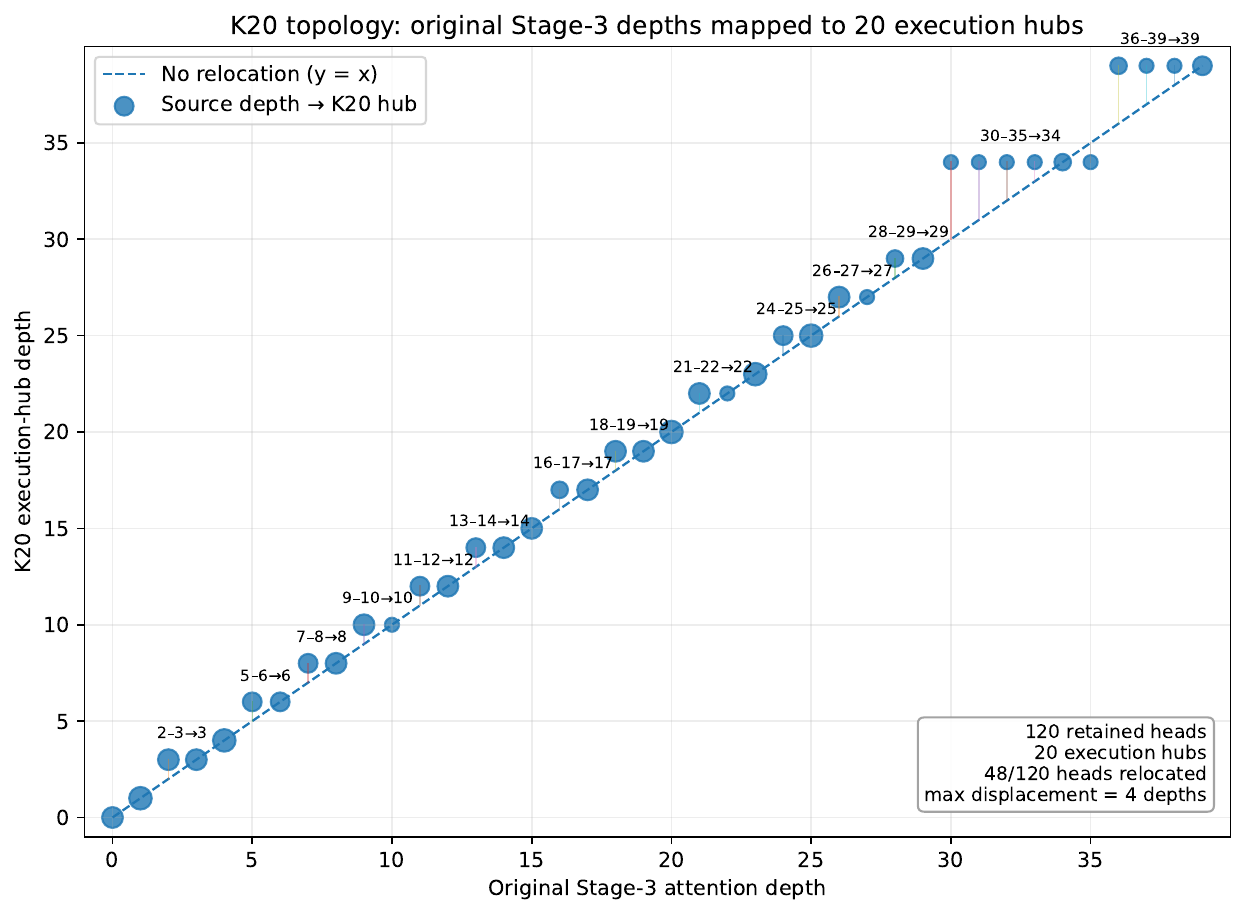}
\caption{Frozen K20 execution topology. Each point maps an original Stage-3 attention depth to the execution depth of its assigned K20 hub. K20 retains all 120 State120 heads while consolidating them into 20 hubs; 48 heads are relocated and the maximum displacement is four original depths.}
\label{fig:supp_topology}
\end{figure}

\begin{figure}[ht]
\centering
\includegraphics[width=0.92\linewidth]{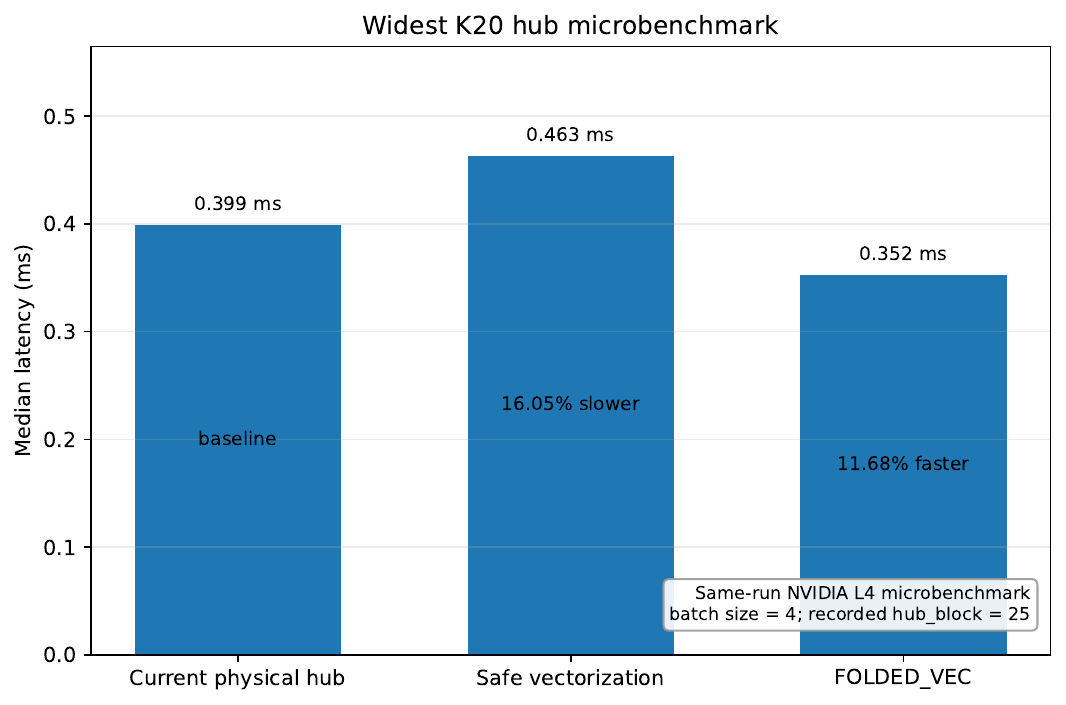}
\caption{Microbenchmark of the widest K20 attention hub on NVIDIA L4 at batch size 4. Direct safe vectorization increases latency, whereas FOLDED\_VEC reduces latency by 11.68\% relative to the current physical hub.}
\label{fig:supp_microbenchmark}
\end{figure}

\end{document}